\documentclass[10pt]{article}

\usepackage{graphicx} 
\usepackage{amsmath}
\usepackage{amssymb}
\usepackage{booktabs}
\usepackage{comment}
\usepackage{multirow}
\usepackage{color}
\usepackage{url}
\usepackage{subcaption}

\newtheorem{remark}{Remark}

\newcommand{\free}{\mathcal{M}}

\newcommand{\cond}{X}
\newcommand{\dmodel}{{d_{\rm model}}}
\newcommand{\nin}{n_{in}}
\newcommand{\nsamp}{N}
\newcommand{\R}{\mathbb{R}}
\newcommand{\norm}[1]{\left\lVert#1\right\rVert}

\newcommand{\E}{\mathbb{E}}
\newcommand{\sys}{S}

\newcommand{\D}{\mathcal{D}}

\title{\LARGE \bf 
Enhanced Transformer architecture for in-context learning of dynamical systems
}
\author{Matteo Rufolo, Dario Piga,   Gabriele Maroni, Marco Forgione 
\thanks{The authors are  with SUPSI, IDSIA, Lugano, 6962, Switzerland. 
		{\tt\small e-mails: name.surname@supsi.ch}} }
\date{October 2024}

\begin{document}

\maketitle

\begin{abstract}
Recently introduced by some of the authors, the in-context identification paradigm aims at estimating, offline and based on synthetic data, a meta-model that describes the behavior of a whole class of systems. Once trained, this meta-model is fed with an observed input/output sequence (\emph{context}) generated by a real system to predict its behavior in a zero-shot learning fashion. In this paper, we enhance the original meta-modeling framework through three key innovations: by formulating the learning task within a probabilistic framework; by managing non-contiguous context and query windows; and by adopting recurrent patching to effectively handle long context sequences. The efficacy of these modifications is demonstrated through a numerical example focusing on the Wiener-Hammerstein system class, highlighting the model’s enhanced performance and scalability. 
\end{abstract}

\section{Introduction}
In system identification (SYSID), the primary objective is to model dynamical systems, leveraging both measured input-output trajectories and prior knowledge of the system's dynamics.
The SYSID setting is thus closely related to supervised  machine learning  (ML), with a specific focus on dynamical systems.  
Due to this similarity, machine-learning techniques have become increasingly popular over the years for estimating dynamical systems through, e.g., neural network architectures~\cite{forgione2021dynonet, bemporad2023training, beintema2023deep} and kernel-based methods~\cite{pillonetto2014kernel,rogers2020application}.

Despite their success, standard supervised learning approaches
do not exploit the insight that could be gained by repeatedly identifying similar systems. While humans improve at solving related tasks, traditional ML (and SYSID) algorithms rely on fixed, predefined procedures that are applied uniformly across different datasets.

The concept of \emph{learning to learn}, or meta-learning, was introduced in the late eighties by the pioneering work~\cite{schmidhuber1987evolutionary} to overcome this limitation and is gaining increasing attention, see~\cite{hospedales2021meta} for a recent survey. In the meta-learning settings, a~\emph{series} of related tasks are presented to an agent which adapts its behavior to act optimally with respect to that class of tasks. In the context of SYSID, the algorithmic agent receives streams of input/output datasets from a class of dynamical systems, and self-tune its behavior to make optimal predictions for those systems~\cite{Chakrabrty2023transformersysid, ZheHal23, achiam2023gpt}.

The meta-model presented by some of the authors in~\cite{forgione2023context} generates multi-step-ahead predictions over a class of dynamical systems. It receives as input a \emph{context} of past input/output samples from a system, together with the future \emph{query} input trajectory, and directly generates an estimate of the corresponding future outputs. 
Training is carried out in a supervised learning manner over a (potentially infinite) stream of \emph{synthetic datasets} obtained by processing randomly-generated input signals through systems randomly sampled from the class of interest. 

To address the meta-modeling task across a broad range of systems, the meta-model must be able to extract relevant knowledge from the context data.
Essentially, the meta-model should function as a SYSID algorithm, learning system-specific models from the context data and leveraging this information to solve the multi-step-ahead simulation task for the given query input.
The peculiar capability of meta-models to behave like learning algorithms is often referred to as \emph{in-context} learning~\cite{brown2020language}.

In~\cite{forgione2023context}, the meta-model was parameterized as an encoder-decoder Transformer, leveraging the strong in-context learning capabilities previously demonstrated by this architecture in the field of natural language processing~\cite{radford2019language}. However, although the Transformers achieved promising results for meta-modeling of dynamical systems, it also has inherent limitations that hinder its potential across common SYSID scenarios. In this paper, we modify the architecture in~\cite{forgione2023context} to address some of the limitations discussed below. 

A well-known critical aspect of Transformers is the computational complexity of their key \emph{attention mechanism}, which grows quadratically with the  sequence length. In~\cite{forgione2023context}, this constrained the context sequence to a few hundred time steps, which is arguably shorter than a typical SYSID dataset. This computational issue has been extensively analyzed in recent literature, with several attempts proposed to mitigate it. Certain approaches aim at approximating attention with simplified mechanisms that achieve comparable results, but with reduced computational complexity~\cite{Beltagy2020LongFormer,Wu2021Autoformer,Zhou2021Informer}. Other contributions adopt a hierarchical processing pattern where sequences are first divided into sub-sequences denoted as \emph{patches}, that are processed individually by a \emph{patching network} which reduces their time dimensionality. The Transformer eventually processes the resulting (shorter) sequence of patch embeddings, instead of the raw samples~\cite{dosovitskiy2021imagepatching,Nie2023timepatching}. In this paper, we address the context length limitation with a patch-based approach inspired by~\cite{Nie2023timepatching}, but utilizing a {Recurrent Neural Network} (RNN)~\cite{Williams1989RNN} as patching network instead of a linear layer.

Another limitations of~\cite{forgione2023context} is that the meta-model's output consists of point estimates, which do not convey information about predictive uncertainty. We address this issue by formulating the learning problem in a probabilistic setting and by estimating the conditional distribution of the query output given the observed context and the query input. 

Finally,~\cite{forgione2023context} assumed contiguous context and query windows. The task was formalized as a time series forecasting problem, where the objective is to guess the future values of the time series, given the past observations and future inputs. However, SYSID generally aims at estimating simulation models, that can describe systems starting from arbitrary initial conditions. To align with the SYSID setting, we modify the task and also provide the first input/output samples of the query segment to the meta-model. These samples convey the initial condition information for the query sequence, thus allowing it to be detached from the context.

The rest of the paper is organized as follows. Section~\ref{sec:prob_descr} describes the problem setting, analyzing in  detail the meta-modeling framework in~\cite{forgione2023context} and the limitations of the architecture introduced in that work. The salient architectural changes introduced in this paper to overcome these limitations are described in~\ref{sec:arch_ext}.
A numerical example is illustrated in Section~\ref{sec:example} to demonstrate the effectiveness of the proposed methodology. Conclusions and the direction for future studies are discussed in the Section~\ref{sec:conc}.

\section{Problem description}
\label{sec:prob_descr}
In~\cite{forgione2023context}, an in-context parametric learner $\free_{\phi}$, referred to as meta-model, has been introduced to describe a  class of dynamical  systems. 
The meta-model $\free_{\phi}$, with tunable parameters $\phi$, 
is trained on a ``meta-dataset".
To define the ``meta-dataset'', two \emph{probability  distributions}, one over dynamical systems  and one over input signals, are introduced. By sampling from these two distributions, it is possible to generate a sequence of systems and inputs that, together with the corresponding outputs, result in ``usual'' input/output datasets. This leads to a potentially infinite \emph{stream} of datasets $\{ \mathcal{D}^{(i)} = (u_{1:N}^{(i)}, y_{1:N}^{(i)}), \, i=1,2,\dots, \infty \}$
with $u_k^{(i)} \in \mathbb{R}^{n_u}$ and $y_k^{(i)} \in \mathbb{R}^{n_y}$, each sampled from the dataset distribution $p(\mathcal{D})$ induced by randomly sampling systems and inputs. The datasets $\mathcal{D}^{(i)}$ are all different, but they are drawn from the same probability distribution, so it is possible to transfer the learned knowledge from one realization to another.

Each dataset realization  $\D^{(i)}$  is split into an initial \emph{context} segment  of length $m$ and a contiguous \emph{query} segment of length $N-m$.
The meta-model $\free_{\phi}$ is trained to reconstruct the query output $y_{m+1:N}^{(i)}$ from the query input $u_{m+1:N}^{(i)}$ and the input/output context  $(u_{1:m}^{(i)}, y_{1:m}^{(i)})$, namely:
\begin{equation}
\label{eq:Nodel_free_sim}
\hat y_{m+1:N}^{(i)} = \free_\phi( u_{m+1:N}^{(i)},  u_{1:m}^{(i)}, y_{1:m}^{(i)}), 
\end{equation}
where $\hat y_{m+1:N}^{(i)}$ represents the estimate of the output $y_{m+1:N}^{(i)}$. 

Training is performed in a supervised manner, by minimizing the regression loss
\begin{equation}
	\label{eq:simulation_objective_samples}
    J = 
    \frac{1}{b}    \sum_{i=1}^b
    \norm{y_{m+1:\nsamp}^{(i)} - \free_\phi (u_{m+1:\nsamp}^{(i)}, u_{1:m}^{(i)}, y_{1:m}^{(i)})}^2
    ,
\end{equation}
where $b$ denotes the minibatch size, namely  the number of  datasets randomly extracted at each iteration of gradient-based optimization. Note that each dataset $\mathcal{D}^{(i)}$ is associated to a different dynamical system $\sys^{(i)}$, with $i=1, \ldots,b$. 

Intuitively, the meta-model $\free_{\phi}$ is expected to ``understand'' the dynamics of system $\sys^{(i)}$ from the context data $(u_{1:m}^{(i)}, y_{1:m}^{(i)})$ and to use this knowledge to generate output predictions $\hat y_{m+1:N}^{(i)}$ in the query segment. Eventually, the trained meta-model $\free_\phi$, will be able to generate multi-step-ahead predictions for a  \emph{new} system  $\sys$ given an input/output sequence $(u_{1:m}, y_{1:m})$ and the query input $u_{m+1:\nsamp}$, without the need of estimating a model for that system. It is worth remarking that, although training  is done offline based on  potentially synthetic data generated in simulation, the trained meta-model is supposed to be applied online to measured data from a real system.

In~\cite{forgione2023context}, an encoder-decoder Transformer architecture~\cite{vaswani2017attention} is used as meta-model. 
The input/output context data $(u_{1:m}, y_{1:m})$ are processed by the encoder, which produces a real-valued sequence $\zeta_{1:m},\, \zeta_i \in \R^{d_{\rm model}}$. Then, the decoder generates output predictions $\hat y_{m+1:N}$ by processing the query input $u_{m+1:N}$ causally in time through masked self-attention, and integrating the encoder's output $\zeta_{1:m}$ through cross-attention. Basically, the encoder output $\zeta_{1:m}$ provides the decoder with the required information about the underlying dynamics, and it could be interpreted as an implicit representation (thus, a model) of the system. For a visual representation of this architecture, see Fig.~2 in~\cite{forgione2023context}.

The contribution in~\cite{forgione2023context} has some  limitations, strictly related with the chosen architecture:
\begin{enumerate}
    \item[L1:] The meta-model only provides point estimates of the future outputs, instead of a probabilistic  distribution that could inform on predictive uncertainty.
    \item[L2:] The context and query segments are required to be contiguous, according to a \emph{time series forecasting} framework. 
    However, in SYSID, one usually seeks \emph{simulation model} that can provide future predictions for arbitrary initial conditions. 
    \item[L3:] The  multi-head attention layers in the Transformer's encoder has a computational complexity that scales  quadratically with respect to the length of the context. This limits the length of the context sequence (roughly, order of magnitude of 1000 samples when training on a single GPU).
\end{enumerate}

In this paper, we aim to overcome these limitations. Specifically: L1 is tackled by formulating the learning  problem in a probabilistic setting; L2 through a  modification of the  architecture that allows the decoder  to process input and output samples previous to the query as initial conditions;  and L3  by dividing the context sequence into \emph{patches}.

\section{Advancing the Transformer Architecture}
\label{sec:arch_ext}
\begin{figure*}[!bt]
\centering
\includegraphics[width=.95\textwidth]{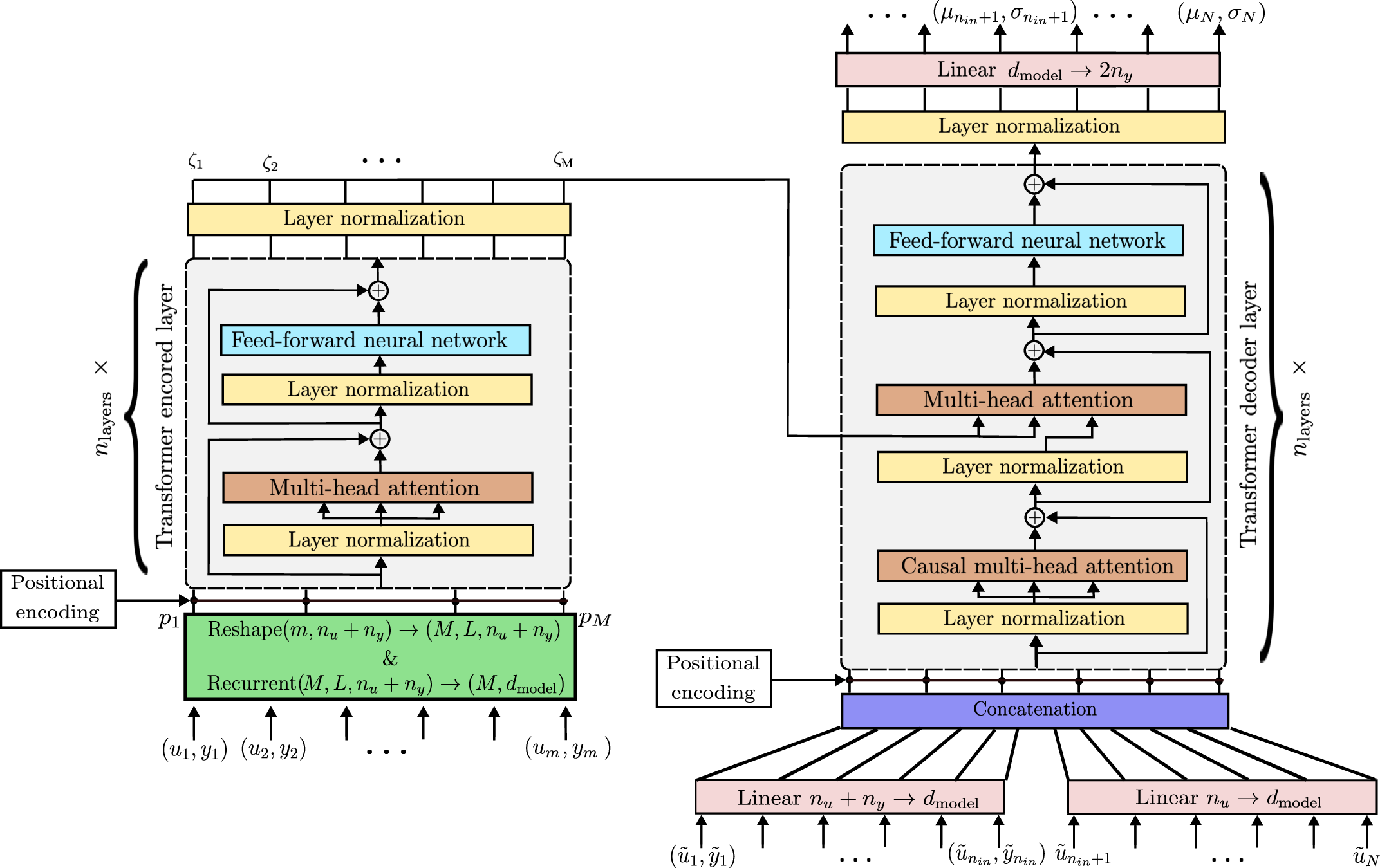}
\caption{Final Transformer architecture to handle probabilistic prediction (top right: decoder's output is a sequence of mean and standard deviation values); non-contiguous context and query (bottom right: initial input/output values of the query fed to the decoder); long context sequences (bottom left: context sequence split into patches and processed by an RNN).}
\label{fig:encoder_decoder_patch}
\end{figure*}

The Transformer architecture in~\cite{forgione2023context} has been extended to address the three limitations previously mentioned. In the next paragraphs, we will systematically address each limitation, detailing the specific modifications made to the original architecture. We anticipate the  final extended architecture  in Fig.~\ref{fig:encoder_decoder_patch}, with the following main changes: 
\begin{itemize}
    \item to address L1, we modify the final output layer to provide both the mean and the standard deviation of the predicted output samples;
    \item to address L2, we introduce an additional layer before the decoder to handle arbitrary initial conditions of the query sequence;
    \item to address L3, we split the context sequence into patches, which are individually processed by a RNN  before being fed into the multi-attention blocks of the Transformer's encode.
\end{itemize}

\subsection{Learning Probability Distributions}

To formulate the learning problem within a probabilistic setting, we introduce the conditional probability distribution $p(y_{m+1:\nsamp} | u_{m+1:\nsamp}, u_{1:m}, y_{1:m})$ of the future output sequence $y_{m+1:N}$, given the query input $u_{m+1:\nsamp}$ and the context data $u_{1:m}, y_{1:m}$. For compactness, we denote  by $\cond$ the conditioning variables $u_{m+1:\nsamp}, u_{1:m}, y_{1:m}$ hereafter.

Our goal is to approximate $p(y_{m+1:\nsamp} | \cond)$ with a parametric distribution $q_{\phi}(y_{m+1:\nsamp} | \cond)$ having free parameters $\phi$. To this end, we introduce the Kullback-Leibler divergence $\text{KL}(p \parallel q_{\phi})$ between $p$ and $q_\phi$:
\begin{equation}
    \label{eq:KL}
  \text{KL}(p \parallel q_\phi)  =  \E_{p({y_{m+1:N}}|\cond)}  \left [  \log \frac{p(y_{m+1:N}| \cond)} {q_{\phi}(y_{m+1:N} | \cond)} \right].
\end{equation}
According to the maximum likelihood estimation principle, the parameters $\phi$ are chosen so as to minimize $E_{p(\cond)} \big [ \text{KL}(p \parallel q_\phi) \big ]$,
namely the expected value (over the conditioning variables $\cond$) of the Kullback-Leibler divergence~\cite{murphy2022probabilistic}. We have:
\begin{multline}
    \label{eq:KL_simplified}
    E_{p(\cond)} \big [ \text{KL}(p \parallel q_\phi) \big ] = E_{p(\D)}  \left [  \log \frac{p(y_{m+1:N}| \cond)} {q_{\phi}(y_{m+1:N} | \cond)} \right] \\
    = \overbrace{E_{p(\D)} [-\log q_{\phi}(y_{m+1:N} | \cond) ]}^{J(\phi)} + K,
\end{multline}
where $J(\phi)$ is the loss function to be minimized w.r.t. $\phi$,  and $K$ is a term does not depend on $\phi$. 
To evaluate $J(\phi)$, the expectation over $p(\D)$, which is in general analytically intractable, is replaced by a Monte Carlo estimate over $b$ sampled datasets: 
\begin{equation}
    \label{eq:J}
     J(\phi) \approx \frac{1}{b} \sum_{i=1}^b -\log q_\phi(y_{m+1:N}^{(i)} | \cond^{(i)}),
\end{equation}
where $y_{m+1:N}^{(i)}$ and $\cond^{(i)}, i=1,\dots,b$ are samples from $p(\D)$. 

In practice, an iterative gradient-based algorithm is applied to minimize $J(\phi)$.
Each optimization iteration involves evaluating \eqref{eq:J} over $b$ newly sampled datasets, computing its gradient with respect to $\phi$, and updating the parameters accordingly. In other words, training is performed in a (minibatch) stochastic gradient descent fashion, where $b$ plays the role of the minibatch size.

In this work, we model the parametric distribution $q_\phi$ as a multivariate Gaussian with diagonal covariance. For the sake of exposition, we consider the single-output case (i.e., $n_y=1$) in the rest of this paragraph. Thus,  $q_\phi$ takes form:
\begin{subequations}
\begin{align}
  q_\phi(y_{m+1:\nsamp} | \cond) &= \prod_{j=m+1}^\nsamp  q_\phi(y_{j} | \cond)\\
  q_\phi(y_{j} | \cond) &= \frac{1}{\sqrt{2\pi \sigma_j^2(\cond, \phi)}}e^{\left(-\frac{(y_j - \mu_j(\cond, \phi))^2}{2\sigma_j^2(\cond, \phi)}\right)},
\end{align}
\end{subequations}
 where the mean and standard deviation vectors $\mu_{m+1:\nsamp}$, $\sigma_{m+1:\nsamp} \in \mathbb{R}^{N-m}$ are functions of $\cond$ and $\phi$. 

These two vectors are the output of the Transformer, when fed with the query input $u_{m+1:N}$ and the input/output context $u_{1:m}, y_{1:m}$ from a system $\sys$:
\begin{equation}
 \label{eq:Nodel_free_sim}
 \mu_{m+1:\nsamp}, \sigma_{m+1:\nsamp} = \free_\phi( u_{m+1:\nsamp},  u_{1:m}, y_{1:m}).
 \end{equation}
Thus, with respect to~\cite{forgione2023context}, we modify the final layer of the decoder to output the two vectors: $\mu_{m+1:N}$ and $\sigma_{m+1
} \in \mathbb{R}^{\nsamp - m}$  rather than providing a single point estimate vector $\hat y_{m+1
:N} \in \mathbb{R}^{\nsamp - m}$, see the upper right block in Fig.~\ref{fig:encoder_decoder_patch}. It is worth remarking that the vector of means $\mu_{m+1:\nsamp}$  corresponds to the point estimate $\hat y_{m+1:N}$ already provided by the previous architecture, while $\sigma_{m+1:\nsamp}$ gives new insight about the meta-model's predictive uncertainty.

\subsection{Handling arbitrary initial conditions}
In order to consider distinct context and query sequences, it is necessary to provide initial conditions along with a query input sequence. In this work, we feed the meta-model with $\nin$ input-output samples preceding the query input as the initial conditions.  
Thus, for supervised learning of this meta-model $\free_{\phi}$,  two input/output sequences are generated from each system $\sys$: a context sequence $(u_{1:m}, y_{1:m})$ and a disjoint query sequence $(\tilde{u}_{1:{\nsamp}}, \tilde{y}_{1:{\nsamp}})$, whose initial samples $(\tilde{u}_{1:{\nin}}, \tilde{y}_{1:{\nin}})$ are taken as initial conditions.

Then, given a query input $\tilde{u}_{\nin+1:N}$, initial conditions $(\tilde{u}_{1:{\nin}}, \tilde{y}_{1:{\nin}})$, and context $(u_{1:m}, y_{1:m})$, the meta-model returns the mean and standard deviation vectors $\mu_{\nin+1:N}$, $\sigma_{\nin+1:N}$ of the probability distribution of the future output sequence $\tilde{y}_{\nin+1:N}$:
\begin{equation}
\label{eq:Nodel_free_sim_CI}
\mu_{\nin+1:N}, \sigma_{\nin+1:N} = \\
\free_\phi(\tilde{u}_{\nin+1:N}, \tilde{u}_{1:{\nin}}, \tilde{y}_{1:{\nin}}, u_{1:m}, y_{1:m}).
\end{equation}

To this end, we modify the Transformer architecture by introducing a linear layer that processes the initial conditions, mapping each time step from $\mathbb{R}^{n_{u} + n_y}$ to $\mathbb{R}^{\dmodel}$. Similarly, the query input is mapped from $\mathbb{R}^{n_{u}}$ to $\mathbb{R}^{\dmodel}$ through a separate linear layer. The corresponding output sequences are then concatenated to form a single sequence of length $N$, where all elements have dimension $\dmodel$. This sequence is combined with positional encoding before being passed into the decoder backbone of the Transformer, as shown in the layers before the decoder in Fig.~\ref{fig:encoder_decoder_patch}.

\begin{remark}
The length $m$ of the input/output context sequence should be sufficiently large to characterize the dynamics of the system. On the other hand, the number $\nin$ of input/output samples used as initial conditions is typically small, of the order of magnitude of the number of dynamical states of the system. \hfill $\blacksquare$
\end{remark}

\subsection{Patching for long context sequences}
\label{subsec:patching}

In order to handle Limitation L3, and thus considering long context sequences, we adopt a patching approach tailored to the meta-learning problem introduced in Section~\ref{sec:prob_descr}, and described in the following paragraphs. 

\begin{figure}[!bt]
\centering
\includegraphics[width=0.8\textwidth]{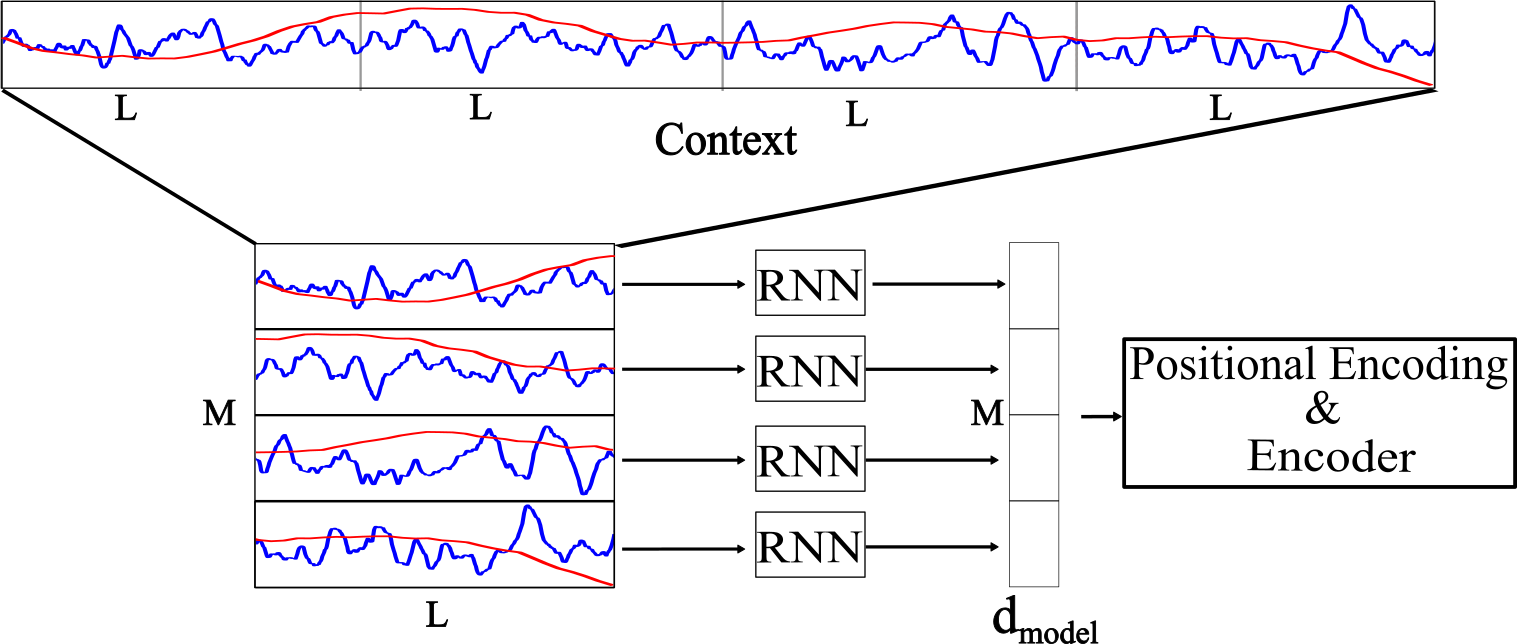}
\caption{Visual representation of the implemented patching approach.  For the sake of visualization, a single-input ($n_u=1$) single-output ($n_y=1$) system is considered. Therefore, the context consists of an input (blu) and output (red) sequence.}
\label{fig:patch_structure}
\end{figure}

The input-output context sequence $(u_{1:m}, y_{1:m})$ is divided into $M$ non-overlapping patches\footnote{For simplicity of exposition, we assume that $m$ is multiple of $M$.}, each of length $L=\frac{m}{M}$ and dimension $n_u+n_y$. The patches are then processed by an RNN which maps each of them into a single vector of dimension $\dmodel$.  The resulting \emph{patch embedding} sequence $p_{1:M}$ is then combined with position encoding to account for the temporal order and then fed as input to the encoder's  backbone (see  Fig.~\ref{fig:patch_structure}). 
The modifications of the architecture with respect to~\cite{forgione2023context} corresponds to the bottom-left blocks in  Fig.~\ref{fig:encoder_decoder_patch}.  Thus, the length of the sequence processed by the multi-head attention mechanism processes is reduced by a factor $L$, significantly decreasing the computational burden by a factor of $O(L^2)$. 

\section{Numerical Example}
\label{sec:example}
The presented meta-model architecture is assessed for in-context learning of dynamical systems belonging to the Wiener-Hammerstein (WH)~\cite{giri2010block} class.
For meta-model learning, the negative log-likelihood~\eqref{eq:J} is minimized over 1 million iterations of the AdamW algorithm~\cite{loshchilov2018adamW}, with minibatch size $b=32$. Training is repeated for different values of the context length $m \in \{400, 800, 16000, 40000\}$.
All computations are performed on a server equipped with an Nvidia RTX 3090 GPU.  For reproducibility of the results, all codes made are available in the GitHub repository~\cite{repo}.

\subsection{System Class and Dataset Distribution}
We generate synthetic datasets from random stable WH systems having structure $G_1\!-\!F\!-\!G_2$, where $F$ is a static non-linearity sandwiched between two Linear Time-Invariant (LTI) dynamical blocks $G_1$ and $G_2$. Systems are drawn from the same distribution considered in~\cite{forgione2023context}. The discrete-time LTI blocks $G_1, G_2$ are randomly chosen with order between $1$ and $10$. The magnitude and phase of their poles are randomly chosen from uniform distributions in the ranges $(0.5, 0.97)$ and $(0,\pi/2)$, respectively. The non-linear block $F$ is a feed-forward neural network with one hidden layer; 32 hidden units; and weights randomly generated from a Gaussian distribution with Kaiming scaling~\cite{he2015delving}.
The input signals fed to the WH systems are drawn from a standard Normal distribution. The WH system output is standardized to have zero mean and unit variance and corrupted by additive white Gaussian noise with standard deviation $\sigma_{\text{noise}} = 0.1$.

\subsection{Transformer architecture}
The chosen Transformer architecture is characterized by: $n_{\rm layers} = 12$ layers; $\dmodel=128$ hidden units; and $n_{\rm heads} = 4$ attention heads, both in the encoder and the decoder. The number of the initial conditions is set to $n_{\rm in} = 10$ and the query length to $N-n_{\rm in}=100$.

As previously mentioned, training is performed for context length $m \in \{400, 800, 16000, 40000\}$.
For the sake of comparison between the proposed patching-based approach and the non-patching method, for the shortest context length $m=400$ the recurrent patching approach is   not implemented and the sequence of embeddings $p_{1:M}$ (with $M=m$) is obtained simply by processing the samples $u_{1:m}, y_{1:m}$ through a linear layer with $n_u + n_y$ inputs and $\dmodel$ outputs, as in the original work~\cite{forgione2023context}.

For all other context lengths, the recurrent patching approach described in Section~\ref{subsec:patching} has been followed with fixed embedding sequence length $M=400$, and patch length $L = m/M$. A vanilla RNN with one hidden layer and $\dmodel=128$ hidden units is used to process the $M$ patches, each of length $L$. 
The last hidden unit of each patch is extracted and further processed by a square linear layer of dimension $\dmodel$,
 and finally combined in the patch embedding sequence $p_{1:M}$ fed to the encoder backbone.\footnote{More in general, the number of hidden units of the RNN could be different from $\dmodel$. In that the case, the final linear layer would not be square.}
 The total number of parameters characterizing the Transformer is independent of the context length, and it is equal to $5.54$ millions in all the cases with patching. When patching is not adopted ($m=400$), the number of parameters is $5.5$ millions, slightly smaller since the RNN layer is not present.  

\subsection{Results}
\begin{figure}[!bt]
\centering
\includegraphics[width=.8\textwidth]{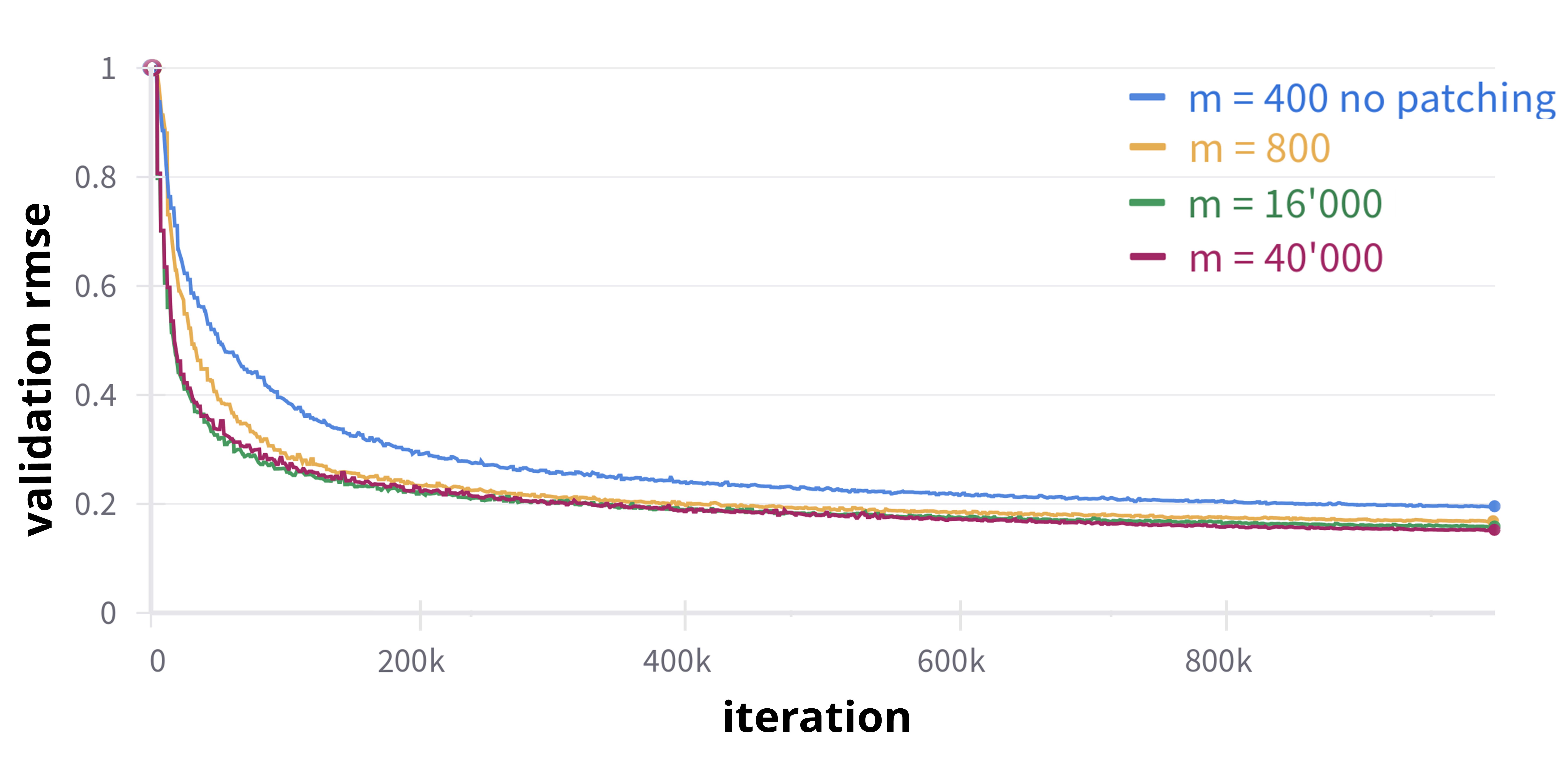}
\caption{Root mean square error in validation vs.  training iteration number  for different context lengths $m$. 
}
\label{fig:Losses}
\end{figure}

The test root mean square error (RMSE), train time, and number of parameters of the different meta-models are reported in Table~\ref{tab:example}. 
Furthermore, the validation RMSE over the iterations of the AdamW optimization algorithm is visualized in Fig.~\ref{fig:Losses}. 

\begin{table}[b!]
    \centering
        \caption{
    Root mean square error in test, train time, and number of parameters of the meta-models for different context lengths $m$.
    }
    \begin{tabular}{*{4}{c}} 
        \toprule 
        $m$ & test RMSE & train time [days] & parameter count [M] \\
        \midrule 
        400 & 0.166 & 1.00  & 5.50 \\
        800 & 0.143 & 1.04  & 5.54 \\
        16000 & 0.128 & 1.67  & 5.54 \\
        40000 & 0.128 & 3.75  & 5.54 \\
        \bottomrule 
    \end{tabular}
    \label{tab:example}
\end{table}

The difference between the achieved RMSE and the noise floor $\sigma_{\rm noise}=0.1$ consistently decreases with the context length  up to $m=16000$. 
We remark that, although it does not result in additional improvements for the benchmark under consideration, the recurrent patching approach allows processing of even longer context sequences. Indeed, we managed to train until convergence a meta-model with context length $m=40000$, thus 100 times larger than the baseline $m=400$ already demonstrated in~\cite{forgione2023context}.
We also note that the $\rm RMSE$ of $0.128$ achieved with $m\geq 16000$ is rather close to the noise floor $\sigma_{\rm noise}=0.1$ induced by the (unpredictable) white Gaussian noise that corrupts the system output.

\begin{figure}[!bt]
\centering
\includegraphics[width=.7\textwidth]{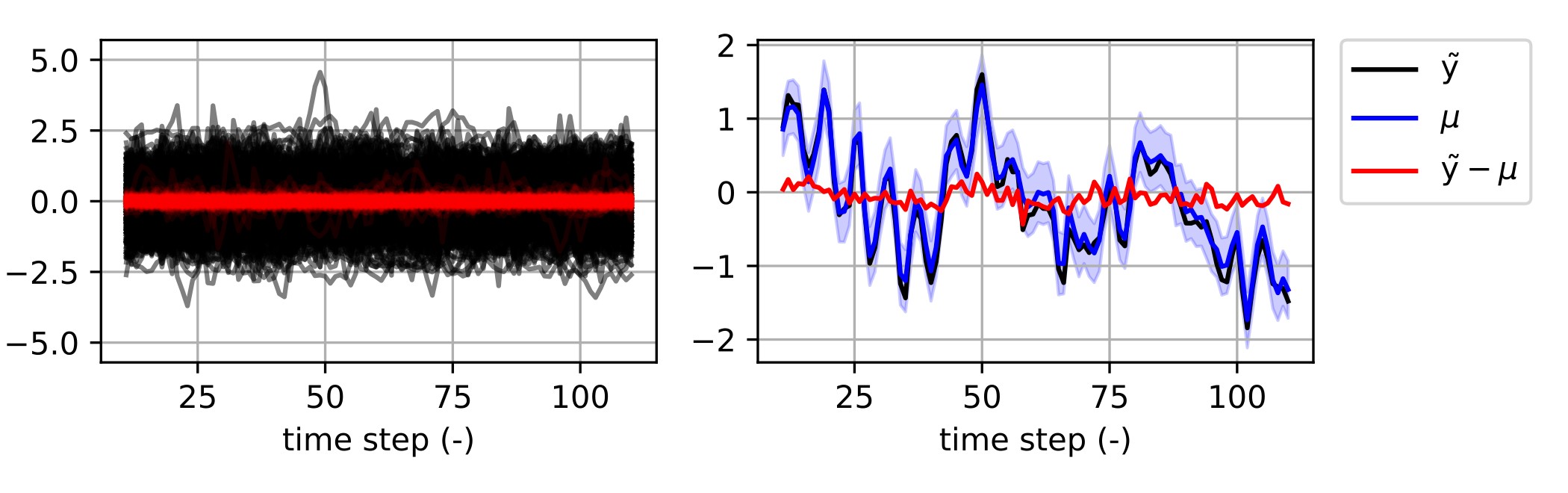}
\caption{Multi-step-ahead simulation  with white noise input of the Transformer trained with $m=16k$. Results on 256 randomly sampled systems superposed (left)
and on a particular system (right). 
Actual output $y$ (black), simulated output mean 
$\mu$ (blue), and simulation error $y - \mu$ (red). The shaded area (light blue) is made by $\pm 3$ standard deviations provided by the meta-model.}
\label{fig:wh_input_noise}
\end{figure}

In Fig.~\ref{fig:wh_input_noise}, the performance of the trained meta-model is visualized. In the left panel, the true WH outputs $\tilde y_{\nin+1:N}$ and the simulation errors $\tilde y_{\nin+1:N} - \mu_{\nin+1:N}$ obtained by the Transformer trained with $m=16000$ over 256 randomly extracted WH systems are shown. In the right panel, for a particular WH realization, the output $y_{\nin+1:N}$, the output mean $\mu_{\nin+1:N}$ predicted by the Transformer, and the prediction error $\tilde y_{\nin+1:N} - \mu_{\nin+1:N}$ are shown. 
Moreover, the $95\%$ credible interval of the prediction computed as  $\mu_{\nin+1:N} \pm 3 \sigma_{\nin+1:N}$ is shown as a shaded area around the nominal Transformer prediction $\mu_{\nin+1:N}$. We observe that the credible intervals effectively capture the actual predictive uncertainty. 

\subsection{Out-of-distribution}
\begin{figure}[!bt]
\centering
\begin{subfigure}[b]{.7\textwidth}
        \centering
        \includegraphics[width=1\textwidth]{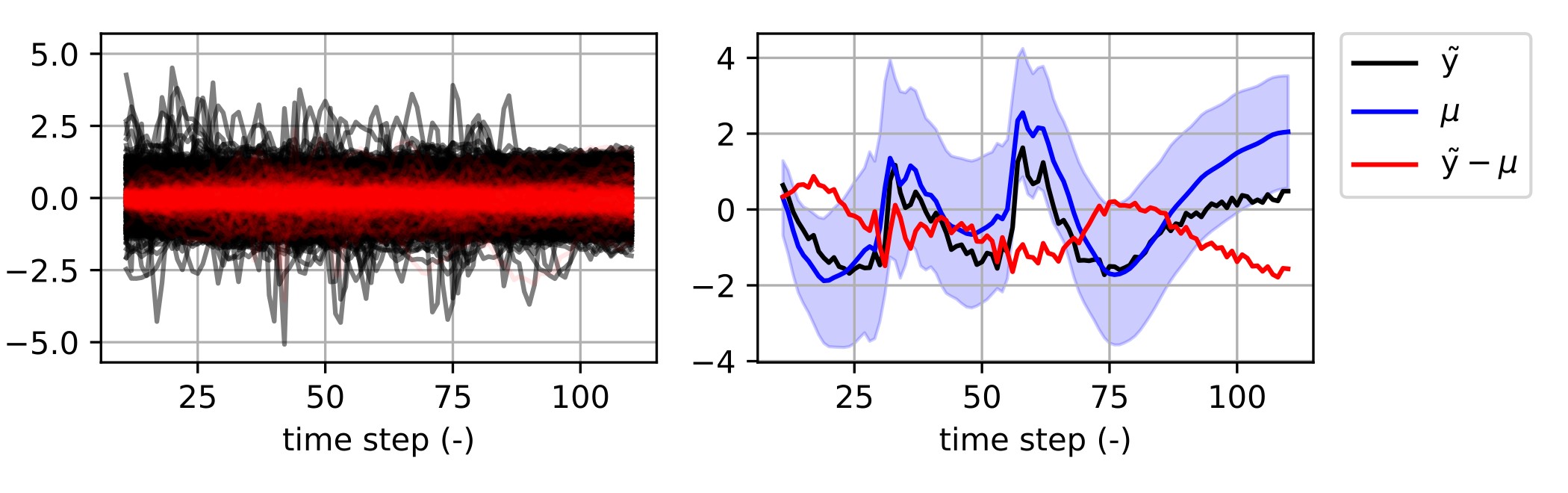}
\caption{PRBS input: simulation results on 256 randomly sampled systems superposed (left)
and on a particular system (right).}
\label{fig:wh_input_binary}
\end{subfigure}

\begin{subfigure}[b]{.7\textwidth}
        \centering
        \includegraphics[width=1\textwidth]{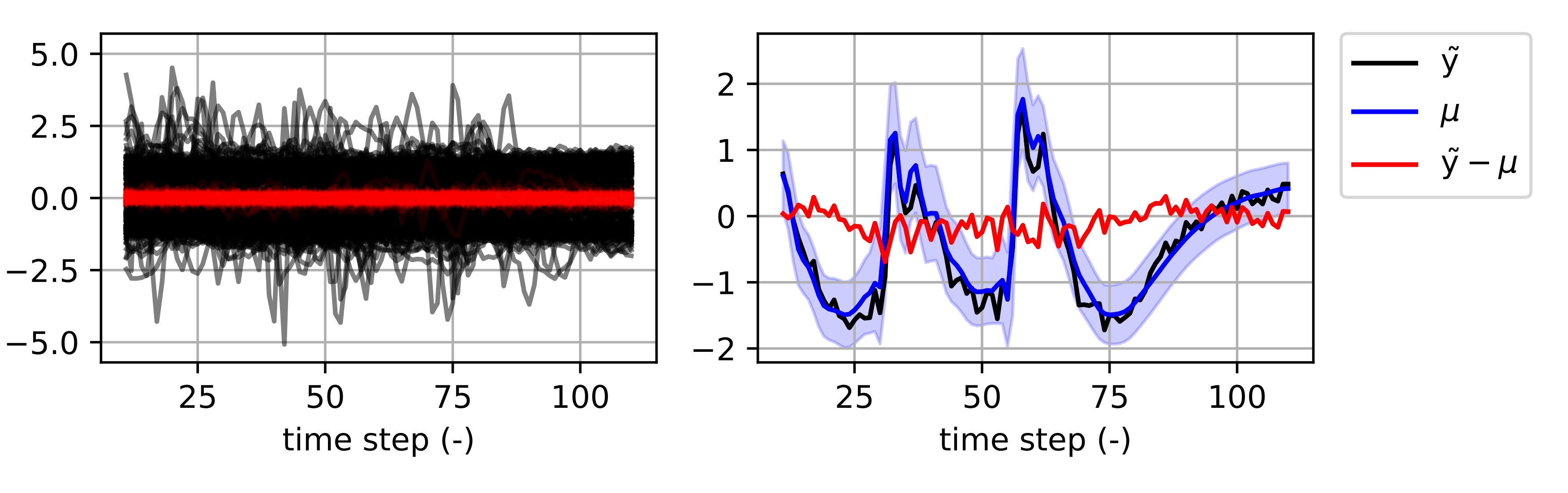}
\caption{Fine-tuned meta-model for PRBS input: simulation results on 256 randomly sampled systems superposed (left)
and on a particular system (right).}
\label{fig:wh_finetuned_input_binary}
\end{subfigure}
\caption{Multi-step-ahead simulation. Meta-model trained with white noise input and tested with PRBS input (top row). Meta-model fine-tuned on PRBS input signals (bottom row). 
Actual output $\tilde{y}$ (black), simulated output mean $\mu$ (blue), and simulation error $\tilde{y} - \mu$ (red). The shaded area (light blue) is made by $\pm$3 standard deviation provided by the meta-model.}
\label{fig:input_binary}
\end{figure}

To assess robustness of the trained meta-model against a shift in the input distribution, we have generated test datasets from $256$ different systems by applying as input  a pseudo random binary signal (PRBS), thus with different characteristics with respect to the input used for meta-model training.  The case of context length $m=16000$ is reported. Output trajectories are visualized 
in Fig~\ref{fig:wh_input_binary}, for all systems (left panel) and for one particular realization (right panel). 
 The average RMSE is 0.3094, about 2.4x larger than the in-distribution case (see Table~\ref{tab:example}).  It is interesting to observe that, as the performance decrease, the estimated uncertainty bands get wider.

Following the approach in~\cite{piga2023adaptation}, we performed a short fine-tuning of the meta-model for PRBS input over 40k iterations, corresponding to approximately 2 hours of training. Results are visualized in Fig.~\ref{fig:wh_finetuned_input_binary} for the same test datasets considered in the previous paragraph. We remark that fine-tuning leads to a significant improvement in the RMSE from 0.3094 to 0.1058 (thus very close to the noise floor $\sigma_{\rm noise}=0.1$), along with a shrinkage of the uncertainty bands.

\section{Conclusions}
\label{sec:conc}

Major modifications to the Transformer-based meta-model architecture introduced in~\cite{forgione2023context} for in-context system identification have been presented. 
Numerical experiments show that using longer context windows results in a substantial accuracy improvement, with the trained meta-model approaching the noise floor. Furthermore, the meta-model maintains reasonable performance in the case of out-of-distribution inputs, with prediction capabilities that can be significantly improved through fast fine-tuning. 

The statistical formulation, besides providing uncertainty quantification for the output predictions, opens up applications in other areas, such as meta-state estimation~\cite{busetto2024}, or in classical system identification tasks where the meta-model can be used to generate synthetic data  as in~\cite{piga2024syntheticdatagenerationidentification}, and the uncertainty can be used to properly weight synthetic samples. This is the subject of ongoing research, along with activities  to further enhance the scalability and computational efficiency of the approach.

\bibliographystyle{IEEEtran} 
\bibliography{root}

\end{document}